\documentclass{article}

\usepackage{amsfonts}
\usepackage{spconf,amsmath,graphicx}

\DeclareMathOperator{\EX}{\mathbb{E}}
\newcommand\numberthis{\addtocounter{equation}{1}\tag{\theequation}}
\usepackage{enumitem}
\setlist{nosep, leftmargin=14pt}

\usepackage{mwe} 
\usepackage{hyperref}

\title{Domain Generalization by Learning from Privileged Medical Imaging Information}
\name{%
\begin{tabular}{@{}c@{}}
Steven Korevaar$^\dag$\sthanks{This research was supported by grants from NVIDIA and utilized an NVIDIA Quadro A6000 for running experiments.\\ \\
This work has been submitted to the IEEE for possible publication. Copyright may be transferred without notice, after which this version may no longer be accessible.
}, 
Ruwan Tennakoon$^\dag$,
Ricky O'Brien$^\dag$,
Dwarikanath Mahapatra$^\ddag$, and \\
Alireza Bab-Hadiashar$^\dag$
\end{tabular}}

\address{$^\dag$RMIT University, Melbourne, Australia.\\
$^\ddag$Inception Institute of AI (IIAI), Abu Dhabi, UAE.
}

\begin{document}
%
\maketitle
\begin{abstract}
Learning the ability to generalize knowledge between similar contexts is particularly important in medical imaging as data distributions can shift substantially from one hospital to another, or even from one machine to another. To strengthen generalization, most state-of-the-art techniques inject knowledge of the data distribution shifts by enforcing constraints on learned features or regularizing parameters. We offer an alternative approach: Learning from Privileged Medical Imaging Information (LPMII). We show that using some privileged information such as tumor shape or location leads to stronger domain generalization ability than current state-of-the-art techniques. This paper demonstrates that by using privileged information to predict the severity of intra-layer retinal fluid in optical coherence tomography scans, the classification accuracy of a deep learning model operating on out-of-distribution data improves from $0.911$ to $0.934$. This paper provides a strong starting point for using privileged information in other medical problems requiring generalization.  
\end{abstract}
\begin{keywords}
Deep learning, medical imaging, generalization, segmentation, classification, privileged information, multi-task learning, optical coherence tomography, OCT.
\end{keywords}
\section{Introduction}
\label{sec:intro}

Deep learning techniques often perform poorly in medical image applications, where data distributions exhibit significant shifts from training data to testing data due to variations in scanners, imaging protocols, etc.
To overcome this challenge of domain shift, domain generalization (DG) techniques have emerged, involving training a model on multiple datasets, each characterized by distinct distributions, with the ultimate goal of developing a model capable of generalizing to unseen test datasets.


Numerous methods for achieving domain generalization have been proposed, and these techniques typically involve incorporating knowledge of data distribution shifts into the learning processes through strategies such as data augmentation (introducing artificial variations in the training data), imposing constraints on learned features, or applying regularization~\cite{zhou_domain_2022, wang_generalizing_2022}. However, extensive empirical evaluations of popular domain generalization algorithms have shown that they consistently fail to outperform ``do-nothing'' baselines like empirical risk minimization (ERM) specifically when fair model selection strategies are employed \cite{gulrajani_search_2020, korevaar_failure_2023}. 

Deep neural network training, especially for classification tasks,  typically focuses on learning the most discriminative feature(s) in the data. However, this can be problematic when these features vary across different contexts leading to poor generalization performance. Humans, on the other hand, tend to learn a broad range of contexts when grasping new concepts, making them more adaptable to different data scenarios. Motivated by the above, \textit{this paper aims to explore whether incorporating extra, task-specific information during training can improve the ability of deep neural networks to adapt to different data contexts}. This additional information, known as ``\textit{privileged information},'' is only available during training and not during testing. In medical imaging, examples of privileged information could include details like tumor shape and location, as well as patient or disease-specific background information found in medical reports. 

While the concept of privileged information has been explored before \cite{VAPNIK2009544}, the application of privileged information for domain generalization has not been previously investigated.
This paper introduces a novel framework designed to integrate privileged information into the learning process for domain generalization tasks. We also assess the viability of this approach through an empirical investigation focused on predicting the severity of intra-layer retinal
fluid in optical coherence tomography (OCT) scans. 



\section{Background}

%

\subsection{Domain Generalization}
There are many techniques that aim to address the domain generalization problem; however, they typically fall under two broad categories: context-free and context-focused. Context-free techniques aim to regularize the learning in some manner to automate the learning of generalizable features. The most common approach is domain feature learning. These methods aim to work across all variations of the domain generalization problem, as they require no additional input, beyond the multiple training domain samples and labels. By comparison, context-focused methods utilize additional knowledge about the context of the problem to achieve generalization. Techniques such as self-supervised learning and data augmentation use handcrafted transformations that can enable a model to be invariant to those forms of transformations; thereby aiding in generalization. For a more detailed breakdown of generalization techniques see~\cite{wang_generalizing_2022}.

\subsection{Learning using Privileged Information}
As with the previous context-focused methods, learning from privileged information \cite{VAPNIK2009544} revolves around the addition of extra information about the dataset that can be used to enhance the performance of a model in some way. Intuitively, the additional contextual information guides the model training process to learn more useful and relevant features for the desired task. 

There are two similar techniques used for addressing generalization issues: self-supervised learning and multi-task learning. Both of these use additional training signals to aid the model in extracting stronger features. Self-supervised learning tasks the model with understanding algorithmically applied variations to data (akin to data augmentation) \cite{bucci_self-supervised_2022, kim_selfreg_2021}, such as classifying a rotation that has been applied to an image. Multi-task learning on the other hand utilises existing and parallel labels to train the model alongside the main task \cite{caruana1997multitask}. 
Our proposed method, using learning from privileged medical imaging information (LPMII), is a targeted subset of multi-task learning; where task-relevant medical knowledge is leveraged to give the model more context about what features may or may not be important for generalization. Although this type of approach has been proposed before, using privileged medical information to improve performance with self-supervised learning \cite{LU2021102094} \cite{pmlr-v126-hu20a} \cite{hervella_2020_joint} \cite{holmberg_self-supervised_2020}, it has been used to improve in-domain (or i.i.d) performance as opposed to targeting generalization across different data distributions.

\section{Methodology}

Domain Generalization aims to learn a robust predictive model $f_\theta : \mathcal{X} \rightarrow \mathcal{Y}$ from $K$ source domain datasets, $\{\mathcal{S}^k \sim P_{XY}^{\mathcal{S}^k} \}_{k=1}^K$, that minimizes the prediction error when applied to an unseen target domain $\mathcal{T} \sim P_{XY}^{\mathcal{T}}$. Here the target data distribution is different from the source domains i.e., $P_{XY}^{\mathcal{T}} \neq  P_{XY}^{\mathcal{S}^K} ~\forall k \in [1\dots K]$. 

We consider a supervised learning scenario where we have multiple source domains. Each source domain has $N_k$ input-target pairs $\left \{ \left (x_i, y_i \right) \right \}_{i=1}^{N_k}$, where the inputs (covariates) are noted by $x_i$ and the targets by $y_i$. Moreover, we assume that each source domain also has privileged information $q_i \in Q$ that is related to the task. Privileged information is only available during training, not during testing. Therefore, each source domain has $N_k$ triplets of the form $\left \{ \left (x_i, q_i, y_i \right) \right \}_{i=1}^{N_k}$.

\begin{figure}[ht]
    \centering
    \includegraphics[width=0.90\linewidth,keepaspectratio]{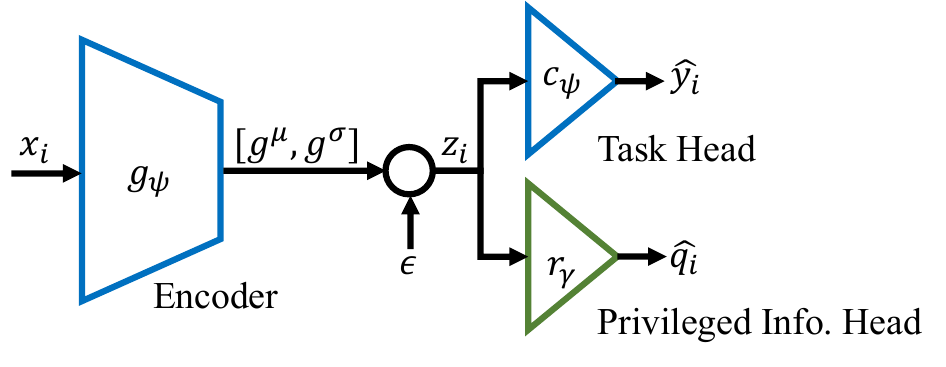}  
    \caption{The model structure of learning from privileged medical information (LPMII). The feature extracting variation encoder used was a ResNet50 structurally.}
    \label{fig:side_info_structure}
\end{figure}

Our proposed method, LPMII, leverages the privileged information to enhance the learning process.
Assume that the model $f_\theta$ consists of a stochastic encoder (or feature extractor) $g_\phi: \mathcal{X} \rightarrow \mathcal{Z}$ and a classifier head $c_\psi: \mathcal{Z} \rightarrow \mathcal{Y}$. Since the privileged information is relevant to the task and provides context, our goal is to capture that in the feature representation, $Z$. 
To do this, we maximize the mutual information between the learned features, $Z$, and the privileged information, $Q$, on the source training data, $S$. The learning objective can be expressed as follows:
\begin{equation}
    \underset{\phi, \psi}{\mathrm{arg~min}} ~ \underset{x,y \sim P_{XY}^\mathcal{S}}{\mathbb{E}} \left[ \mathcal{L}_{ce}\left (y, c_\psi \left (g_\phi(x) \right) \right ) \right ] - \alpha I(Q, {Z})
\end{equation}
Here, the cross-entropy loss $\mathcal{L}_{ce}(\cdot)$ measures the prediction error, while the mutual information $I(Q, Z)$ encourages the features to contain the privileged information. The scalar $\alpha$ controls the trade-off between them. We can compute the mutual information as:
\begin{align*}
\mathrm{I}(Z;Q) &{} = \mathrm{H}(Q) - \mathrm{H}(Q \mid Z) \\
 &{} = \mathrm{H}(Q) + \int_{q,z}  p(q,z) \log{{p(q\mid z)}}.
\end{align*}
By ignoring the constant term $H(Q)$ and using a neural network $r_\gamma(q\mid z)$ to approximate the intractable $p(q\mid z)$, we can derive the following variational lower bound for $\mathrm{I}(Z;Q)$:
\begin{align*}
\mathrm{I}(Z;Q) 
&{} \geq \EX_{x_i, q_i \sim P_{XQ}^\mathcal{S}} \left [ \EX_{z\sim P_{z\mid x}} \left [ \log{{r_\gamma(q_i\mid z)}}  \right ] \right ] \numberthis \label{eqn:final3}
\end{align*}
Using a reparametrization scheme and defining $g_\phi$ as a combination of a deterministic encoder, which produces $\left [g^\mu, g^\sigma \right]$, and a Gaussian random variable $\epsilon \sim p(\epsilon) $ i.e., $g_{\phi,\epsilon}(x) = g^\mu(x) + \epsilon \odot g^\sigma (x)$, we can rewrite the lower bound as:
\begin{align*}
    \mathrm{I}(Z;Q) &{} \geq \EX_{x_i, q_i \sim P_{XQ}^\mathcal{S}}  \left [ \mathbb{E}_{p(\epsilon) } \left [ \log{{r_\gamma(q_i\mid g_{\phi,\epsilon}(x))}}  \right ] \right ]\\
    &{} \geq -\EX_{x_i, q_i \sim P_{XQ}^\mathcal{S}}  \left [ \mathbb{E}_{p(\epsilon) } \left [ \mathcal{L}_{ce}\left( q_i, r_\gamma(g_\phi(x)\right)  \right ] \right ]\numberthis \label{eqn:final4}
\end{align*}
If the privileged information takes the form of a regression task, the cross-entropy loss can then be replaced by an appropriate loss function such as the mean squared error. A similar lower bound has been derived in~\cite{chechik2002extracting}. The final objective function is:
\begin{align*}
    \underset{\phi, \psi, \gamma}{\mathrm{arg~min}} ~ \underset{x,y,q \sim P_{XYQ}^\mathcal{S}}{\mathbb{E}}  & [  \mathcal{L}_{ce}\left (y, c_\psi \left (g_\phi(x) \right) \right )  \\
    & + \alpha  
 \mathbb{E}_{p(\epsilon) } \left [ \mathcal{L}_{ce}\left( q, r_\gamma(g_\phi(x)\right)  \right ]    ]
\end{align*}
The overall block diagram is depicted in Figure \ref{fig:side_info_structure}, where we use a ResNet-50 as the encoder, $g_\phi$, and two MLP classifiers to represent $c_\psi$ and $r_\gamma$.

\begin{figure}[htp]
    \centering
    \includegraphics[width=0.95\linewidth,clip,keepaspectratio]{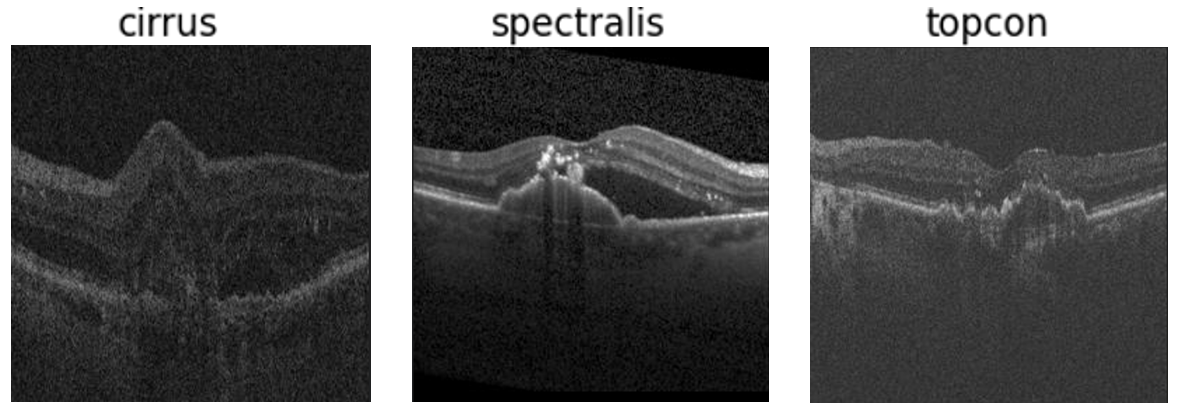}
    \caption{A sample slice from a scan from each manufacturer (Cirrus, Spectralis, and Topcon) within the Retouch dataset. While the underlying structural information is similar, the noise patterns are significantly different leading to domain shift. }
    \label{oct_samples}
\end{figure}

\noindent \textbf{Experimental setup:} To validate this approach a dataset with both multiple domains and adequate additional information is required. The Retinal Optical Coherence Tomography Fluid Challenge (Retouch) OCT \cite{bogunovic_retouch_2019} dataset has both required components. This dataset contains three sets (24, 23, and 21 scans each with 100 slices per scan approximately) of 3D scans of retinas captured using scanners manufactured by three different vendors, each with segmentation masks that indicate the presence of retinal fluids (i,e., Intraretinal fluid, Subretinal fluid, and Pigment Epithelial Detachment). Each 3D scan contains approximately 100 slices, which were separated and resized to $224$x$224$ for training on the 2D ResNet50 network. Examples of the data can be seen in figure \ref{oct_samples}. In this problem, we focus on the classification of the presence of each type of inter-layer retinal fluid for each slice independently. 

For the privileged information source, we derived several ways to extract information from the provided fluid segmentation masks.
Learning from privileged information requires a careful balance between the informativeness and the availability of the data. As such, we use the segmentation masks only to create simpler labels that expert radiologists could easily produce when collecting data. We derived two forms of side information that can be used for this task: (1) Total mass of the fluid build-up, and (2) ``severity'' score.


The most intuitive information that could be extracted from segmentation masks is the mass of the fluid buildup, $q_i^{(mass,l)}$. This was calculated by summing the positive pixels in the segmentation mask, $M_i$, for a given fluid type, $l$. 

\begin{equation*}
q_i^{(mass,l)} = \sum^{H,W}_{h=1,w=1}M_{h,w}^{(l)}
\end{equation*}

Here, $H, W$ are the height and the width of a B-scan. The total mass can be affected by minor errors in the annotation process. In many medical imaging tasks, we observe inter and intra-annotator variations. Additionally, this mass calculation is highly dependent on the resolution of the scan. To address this, a severity estimation was used as well. This severity metric was designed to map the total mass to a scale of 0 to 5 indicating the severity of the fluid build-up, independent of the resolution of the scan. This was achieved by dividing the mass by the total number of pixels in the scan, giving a proportion of how much of the scan is fluid or non-fluid. Given the actual range of values in the data is lower, with the largest amount of normalized fluid being approximately 0.2, the proportion was multiplied by 5 to fully occupy the range of values between 0 and 1. An additional multiplication is done to scale the values to the desired number of severity categories, $N_{sc}$: in our experiments we used $N_{sc} = 5$ to approximate the 0 to 5 severity rating system. This final value was then passed through the ceiling function to discretize the values to integers.

\begin{equation*}
q^{(severity,l)} = \mathrm{ceiling}\left (\frac{5 \cdot N_{sc}}{H \cdot W}\sum^{H,W}_{h=1,w=1} M_{h,w}^{(l)} \right)
\end{equation*}



\begin{figure}[ht]
    \centering
    \includegraphics[width=0.9\linewidth,keepaspectratio]{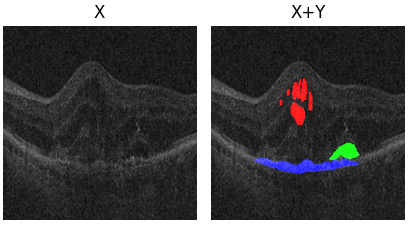}  
    \caption{Visual representation of the two types of privileged information in the Retouch OCT. X) A single slice of a retinal OCT scan. X+Y) The segmentation mask overlaid on the original scan.}
    \label{fig:lpmii_Retouch_centroids}
\end{figure}

The intuition for these metrics was to introduce some correlating information about the fluid. To be able to predict the mass or severity of the fluid build-up the model must be able to accurately identify what sections of the scan are fluid or not, and thus cannot rely on any other correlating factors. Figure \ref{fig:lpmii_Retouch_centroids} shows an example scan along with the segmentation mask.

\section{Results and Discussion}

In this section, we present and discuss the results for the several different variations on the learning from privileged medical information training paradigm. Our method, LPMII was implemented in PyTorch and the code is available at \href{https://github.com/}{Github (available upon publication)}. There are two variations that we will be comparing to each other plus the baseline method, ERM, which has no additional domain generalization methods applied. Each method variation was trained in a hold-one-out fashion, training on two domains and testing on the third, with each domain acting as the test set three times. The out-of-distribution test domain classification accuracies were then averaged together to display the final result for each method.

\begin{table}[ht]
    \small
     \centering
    \caption{Average top-1 classification accuracy for different sources of privileged information.}
    \setlength{\tabcolsep}{3pt}
    \begin{tabular}{r|c}
    \hline
     \textbf{Privileged Information} & \textbf{Average Top-1 Accuracy} \\
    \hline
    \hline
    
    None (ERM)     & 0.911 $\pm$ 0.01\\
    Fluid Mass Regression    & 0.902 $\pm$ 0.02\\
    Severity Prediction    & \textbf{0.934} $\pm$ \textbf{0.01}\\
    
    \hline
    \end{tabular}%
    
    \label{tab:lpmi_Retouch_results}
\end{table}

Additionally, we compare the performance of learning from privileged medical imaging information (LPMII) to seven methods from the DomainBed framework, which represents a wide variety of state-of-the-art domain generalization techniques and mechanisms. The methods chosen are Invariant Risk Minimisation (IRM) \cite{arjovsky_invariant_2020}, Maximum Mean Discrepancy (MMD) \cite{li_domain_2018}, Domain Adversarial Neural Networks (DANN) \cite{ganin_domain-adversarial_2016}, Gradient Matching for Domain Generalization (Fish) \cite{shi_gradient_2021}, Inter-domain Mixup (Mixup) \cite{yan_improve_2020}, Style Agnostic Network (SagNet) \cite{nam_reducing_2021}, and Self-supervised Contrastive Regularization (SelfReg) \cite{kim_selfreg_2021}. The DomainBed framework provides a platform for fare comparison of domain generalization techniques that also include proper model selection methodology \cite{gulrajani_search_2020}.

\begin{table}[ht]
\small
    \centering
    \caption{Average top-1 class-weighted classification accuracy on out-of-distribution data of domain generalization algorithms.}
    \setlength{\tabcolsep}{3pt}
    \begin{tabular}{r|c}
    \hline
    
    \textbf{Method} & \textbf{Average Top-1 Accuracy} \\
    \hline \hline
    ERM & 0.911 $\pm$ 0.01 \\
    IRM \cite{arjovsky_invariant_2020} & 0.896 $\pm$ 0.01 \\
    MMD \cite{li_domain_2018} & 0.907 $\pm$ 0.02 \\
    DANN \cite{ganin_domain-adversarial_2016} & 0.880 $\pm$ 0.02 \\
    Fish \cite{shi_gradient_2021} & 0.911 $\pm$ 0.01 \\
    Mixup \cite{yan_improve_2020} & 0.896 $\pm$ 0.01 \\
    SagNet \cite{nam_reducing_2021} & 0.899 $\pm$ 0.01 \\
    SelfReg \cite{kim_selfreg_2021} & 0.893 $\pm$ 0.01 \\
    \hline
    LPMII & \textbf{0.934} $\pm$ \textbf{0.01} \\
    
    \hline
    \end{tabular}
    \label{OOD_acc_medical_tab}
\end{table}

The above table (table \ref{tab:lpmi_Retouch_results}) shows that the severity prediction task manages to improve the out-of-distribution accuracy by approximately $0.02$ for the Retouch dataset. While this improvement is marginal, it is the first method that has led to a noticeable and consistent improvement in accuracy. Unfortunately, the other privileged information task does not fare as strongly. 

Fluid mass regression performs roughly equivalently to ERM though slightly worse, and with a larger margin of error. Most likely this is due to the network suffering from the requirement to output accurate values above the bounds of 0 to 1, where neural networks typically perform the best. 

While one of the side information methods failed to improve upon ERM, achieving an improvement with the severity prediction indicates that training a model with additional information about the ailment that needs to be classified can be successful in improving generalization performance.

\section{Conclusion}

This article explores the novel use case of improving domain generalization by learning from privileged medical imaging information. This method was designed to introduce a stronger training signal that informs the model about what types of features are more likely to be useful for the main task that will be invariant to domain shifts. We show that by training a model to predict a simple and fast to obtain severity indicator for a retinal layer fluid detection task (from the Retouch dataset) the out-of-distribution binary classification accuracy of the model was increased from $0.911$ to $0.934$. Even though this is a strong preliminary result, it is likely that more informative sources of privileged information could lead to even greater improvements in performance. As such, significantly more testing on learning from privileged information for domain generalization must be done on more challenging medical imaging tasks on a wider variety of different imaging modalities and ailments, with more sources of privileged information to fully explore the potential of this technique.

The most prominent issue standing in the way of this method's wide use is the availability of privileged information. For natural imaging, capturing additional data describing the images is rarely done as a matter of course. Likewise, the types of additional information required to provide an adequate training signal may not be as straightforward. For medical data, there is almost always additional information available in the form of medical reports. These reports often contain highly relevant and useful information pertaining to the issue that is being examined in the scan, as such medical imaging is a prime use case for this technique. 



\bibliographystyle{IEEEbib}
\bibliography{strings,refs}

\begin{thebibliography}{10}

\bibitem{zhou_domain_2022}
Kaiyang Zhou, Ziwei Liu, Yu~Qiao, Tao Xiang, and Chen~Change Loy,
\newblock ``Domain generalization: A survey,''
\newblock {\em IEEE Trans. Pattern Anal. Mach. Intell.}, vol. 45, no. 4, pp.
  4396–4415, apr 2023.

\bibitem{wang_generalizing_2022}
J.~Wang, C.~Lan, C.~Liu, Y.~Ouyang, T.~Qin, W.~Lu, Y.~Chen, W.~Zeng, and P.~S.
  Yu,
\newblock ``Generalizing to unseen domains: A survey on domain
  generalization,''
\newblock {\em IEEE Transactions on Knowledge and Data Engineering}, vol. 35,
  no. 08, pp. 8052--8072, aug 2023.

\bibitem{gulrajani_search_2020}
Ishaan Gulrajani and David Lopez-Paz,
\newblock ``In {Search} of {Lost} {Domain} {Generalization},'' July 2020,
\newblock arXiv:2007.01434 [cs, stat].

\bibitem{korevaar_failure_2023}
Steven Korevaar, Ruwan Tennakoon, and Alireza Bab-Hadiashar,
\newblock ``Failure to {Achieve} {Domain} {Invariance} {With} {Domain}
  {Generalization} {Algorithms}: {An} {Analysis} in {Medical} {Imaging},''
\newblock {\em IEEE Access}, vol. 11, pp. 39351--39372, 2023.

\bibitem{VAPNIK2009544}
Vladimir Vapnik and Akshay Vashist,
\newblock ``A new learning paradigm: Learning using privileged information,''
\newblock {\em Neural Networks}, vol. 22, no. 5, pp. 544--557, 2009,
\newblock Advances in Neural Networks Research: IJCNN2009.

\bibitem{bucci_self-supervised_2022}
Silvia Bucci, Antonio D’Innocente, Yujun Liao, Fabio~M. Carlucci, Barbara
  Caputo, and Tatiana Tommasi,
\newblock ``Self-{Supervised} {Learning} {Across} {Domains},''
\newblock {\em IEEE Transactions on Pattern Analysis and Machine Intelligence},
  vol. 44, no. 9, pp. 5516--5528, Sept. 2022.

\bibitem{kim_selfreg_2021}
Daehee Kim, Youngjun Yoo, Seunghyun Park, Jinkyu Kim, and Jaekoo Lee,
\newblock ``Selfreg: Self-supervised contrastive regularization for domain
  generalization,''
\newblock in {\em Proceedings of the IEEE/CVF International Conference on
  Computer Vision}, 2021, pp. 9619--9628.

\bibitem{caruana1997multitask}
Rich Caruana,
\newblock ``Multitask learning,''
\newblock {\em Machine learning}, vol. 28, pp. 41--75, 1997.

\bibitem{LU2021102094}
Qi~Lu, Yuxing Li, and Chuyang Ye,
\newblock ``Volumetric white matter tract segmentation with nested
  self-supervised learning using sequential pretext tasks,''
\newblock {\em Medical Image Analysis}, vol. 72, pp. 102094, 2021.

\bibitem{pmlr-v126-hu20a}
Szu-Yen Hu, Shuhang Wang, et~al.,
\newblock ``Self-supervised pretraining with dicom metadata in ultrasound
  imaging,''
\newblock in {\em Proceedings of the 5th Machine Learning for Healthcare
  Conference}. 07--08 Aug 2020, vol. 126, pp. 732--749, PMLR.

\bibitem{hervella_2020_joint}
Alvaro~S. Hervella, Lucia Ramos, Jose Rouco, Jorge Novo, and Marcos Ortega,
\newblock ``Multi-modal self-supervised pre-training for joint optic disc and
  cup segmentation in eye fundus images,''
\newblock in {\em ICASSP 2020 - 2020 IEEE International Conference on
  Acoustics, Speech and Signal Processing (ICASSP)}, 2020, pp. 961--965.

\bibitem{holmberg_self-supervised_2020}
Olle~G. Holmberg, Niklas~D. Köhler, et~al.,
\newblock ``Self-supervised retinal thickness prediction enables deep learning
  from unlabelled data to boost classification of diabetic retinopathy,''
\newblock {\em Nature Machine Intelligence}, vol. 2, no. 11, pp. 719--726, Nov.
  2020.

\bibitem{chechik2002extracting}
Gal Chechik and Naftali Tishby,
\newblock ``Extracting relevant structures with side information,''
\newblock {\em Advances in Neural Information Processing Systems}, vol. 15,
  2002.

\bibitem{bogunovic_retouch_2019}
H.~Bogunović, F.~Venhuizen, et~al.,
\newblock ``{RETOUCH}: {The} {Retinal} {OCT} {Fluid} {Detection} and
  {Segmentation} {Benchmark} and {Challenge},''
\newblock {\em IEEE Transactions on Medical Imaging}, vol. 38, no. 8, pp.
  1858--1874, Aug. 2019.

\bibitem{arjovsky_invariant_2020}
Martin Arjovsky, Léon Bottou, Ishaan Gulrajani, and David Lopez-Paz,
\newblock ``Invariant {Risk} {Minimization},'' Mar. 2020,
\newblock arXiv:1907.02893 [cs, stat].

\bibitem{li_domain_2018}
Haoliang Li, Sinno~Jialin Pan, Shiqi Wang, and Alex~C. Kot,
\newblock ``Domain {Generalization} with {Adversarial} {Feature} {Learning},''
\newblock in {\em 2018 {IEEE}/{CVF} {Conference} on {Computer} {Vision} and
  {Pattern} {Recognition}}. June 2018, pp. 5400--5409, IEEE.

\bibitem{ganin_domain-adversarial_2016}
Yaroslav Ganin, Evgeniya Ustinova, et~al.,
\newblock ``Domain-adversarial training of neural networks,''
\newblock {\em The journal of machine learning research}, vol. 17, no. 1, pp.
  2096--2030, 2016.

\bibitem{shi_gradient_2021}
Yuge Shi, Jeffrey Seely, Philip H.~S. Torr, N.~Siddharth, Awni Hannun, Nicolas
  Usunier, and Gabriel Synnaeve,
\newblock ``Gradient {Matching} for {Domain} {Generalization},'' July 2021,
\newblock arXiv:2104.09937 [cs, stat].

\bibitem{yan_improve_2020}
Shen Yan, Huan Song, Nanxiang Li, Lincan Zou, and Liu Ren,
\newblock ``Improve {Unsupervised} {Domain} {Adaptation} with {Mixup}
  {Training},'' Jan. 2020,
\newblock arXiv:2001.00677 [cs, stat].

\bibitem{nam_reducing_2021}
Hyeonseob Nam, HyunJae Lee, Jongchan Park, Wonjun Yoon, and Donggeun Yoo,
\newblock ``Reducing domain gap by reducing style bias,''
\newblock in {\em Proceedings of the IEEE/CVF Conference on Computer Vision and
  Pattern Recognition}, 2021, pp. 8690--8699.

\end{thebibliography}

\end{document}